\documentclass{article}
\usepackage[hyperref]{acl2021}
\usepackage{times}
\usepackage{latexsym}
\usepackage{graphicx}
\graphicspath{images/}

\usepackage{amsmath}
\usepackage{amsfonts}
\usepackage{amsthm}
\usepackage{amssymb}
\usepackage{centernot}
\usepackage{booktabs}
\usepackage{adjustbox}
\usepackage{colortbl}
\usepackage{multirow}
\usepackage{float}
\usepackage{pifont}
\DeclareMathOperator*{\argmax}{argmax}

\definecolor{GPTBlue}{HTML}{5DADE2}
\definecolor{t0green}{HTML}{5bbd4a}
\definecolor{t5orange}{HTML}{ffa500}
\definecolor{t03b}{HTML}{5bbd4a}
\definecolor{t53b}{HTML}{FFA500}
\usepackage{microtype}

\aclfinalcopy 

\newcommand{\curtask}{\mathcal{T}_{\textrm{o}}}
\newcommand{\curchoices}{\mathcal{C}_{\textrm{o}}}
\newcommand{\difftask}{\mathcal{T}_{\textrm{d}}}
\newcommand{\diffprompt}{\mathcal{P}_{\textrm{d}}}

\title{Evaluating Prompts Across Multiple Choice Tasks In a Zero-Shot Setting}

\author{Gabriel Orlanski\\
  \texttt{go533@nyu.edu} \\}

\date{}

\begin{document}
\maketitle
\begin{abstract}
Large language models have shown that impressive zero-shot performance can be achieved through natural language prompts \cite{Radford2019LanguageMA,brown2020language,sanh2021multitask}. Creating an effective prompt, however, requires significant trial and error. That \textit{prompts} the question: how do the qualities of a prompt effects its performance? To this end, we collect and standardize prompts from a diverse range of tasks for use with tasks they were not designed for. We then evaluate these prompts across fixed multiple choice datasets for a quantitative analysis of how certain attributes of a prompt affect performance. We find that including the choices and using prompts not used during pre-training provide significant improvements. All experiments and code can be found \href{https://github.com/gabeorlanski/zero-shot-cross-task}{https://github.com/gabeorlanski/zero-shot-cross-task}.
\end{abstract}

\section{Introduction}
\noindent Recent work has shown that using a natural language (NL) prompt with pre-trained language models (LM) significantly improves performance in few-shot and zero-shot settings \cite{brown2020language,schick-schutze-2021-just}, to the point where they can be worth 100s of data points \cite{Scao2021HowMD}. Further, T5 \cite{raffel-etal-2020} showed that simple prompts and reformulating NLP tasks as text-to-text performs well on a wide range of tasks. Recent models such as FLAN \cite{wei2021finetuned} and T0 \cite{sanh2021multitask} demonstrate that multi-task training a large LM with prompts results in improved zero-shot performance on a wide range of tasks. However, manually designing a prompt is a non-trivial task due to the trial-and-error nature of the task \cite{jiang2020know,shin2021constrained}. Some works focus on ``prompt programming''-- best practices for designing prompts \cite{reynolds2021prompt,liu2021makes}. While others have looked towards continuous ``soft-prompts''-- random vectors added to the input sequence, which are then fine-tuned \cite{zhong2021factual,qin2021learning}. However, recent work has revealed that these prompts are susceptible to minor perturbations \cite{Mishra2021CrossTaskGV}. 

Motivated by this, we aim to conduct a quantitative analysis of what affects a prompt's performance. We evaluate T0-3B \cite{sanh2021multitask} on generalized prompts from a wide range of tasks with eight datasets to provide a quantitative analysis of how the qualitative aspects of a prompt effect its performance. 

We collected 95 prompts across 20 tasks and evaluated each on eight datasets. We find that using a prompt performs better for every evaluation task than not using a prompt. Further, we find that the set of prompts with the highest performance is not the ones designed for the specific task for seven of the eight datasets. In our ablations, we find that adding a small amount of task-specific NL to the generalized prompt increases performance by a median of 4.65\%.
Finally, we find that prompts with the choices present outperform those that do not and that presenting the options as multiple distinct choices further improves the results. Additionally, the longer a prompt is, the worse it performs. 

\section{Methodology}\label{sec:methods_methodology}
\begin{figure}[t]
    \centering
    \includegraphics[width=.95\linewidth,keepaspectratio]{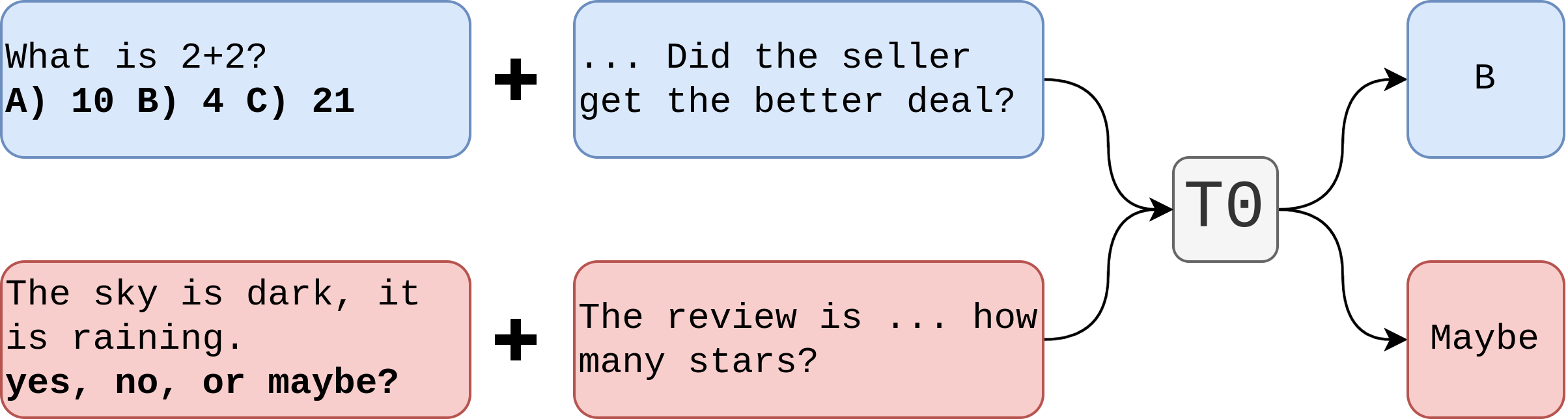}
    \caption{Overview of the approach. For a given example from a dataset (leftmost box) we use a prompt from a different task to perform zero shot predictions with T0-3B \cite{sanh2021multitask}. The \textbf{bolded text} in the example represents its choices.}
    \label{fig:overview}
\end{figure}
Our overall approach is detailed in \autoref{fig:overview}. Given a downstream multiple-choice task $\curtask = \{(\mathbf{x}_1,\mathbf{y}_1),\ldots, (\mathbf{x}_n,\mathbf{y}_n)\}$, evaluate how well T0 \cite{sanh2021multitask} performs when using a prompt $\diffprompt$ designed for a different task $\difftask$. 
\subsection{Fixed Choice Tasks}\label{sec:methods_fixedchoicetasks}
For the purposes of this paper, we limit the scope of the tasks we look at to be only multiple choice tasks whose choices are constant across all examples. Thus, for $\curtask = \{(\mathbf{x}_1,\mathbf{y}_1),\ldots, (\mathbf{x}_n,\mathbf{y}_n)\}$, every $\mathbf{y}_i \in \curchoices$ where $\curchoices = \{\mathcal{C}_1,\ldots ,\mathcal{C}_c\}$ with lengths $\ell=\{\ell_1,\ldots,\ell_c\}$. To make a prediction for the example point $\mathbf{x}_i$, we follow \citet{Holtzman2021SurfaceFC} and use rank scoring: taking the choice with the highest probability as defined by
\begin{equation}\label{eq:predprob}
  \argmax_{\mathcal{C}_j\in\curchoices} \prod_{k=1}^{\ell_j}P(\mathcal{C}_j^k| \mathbf{x}_i, \mathcal{C}_j^1\ldots \mathcal{C}_j^{k-1})
\end{equation}
However, the lengths in $\ell$ are not guaranteed to be the same and thus \autoref{eq:predprob} will unintentionally penalize longer choices \cite{brown2020language,Holtzman2021SurfaceFC}. We thus follow prior work and take the choice with the highest Average Log-Likelihood
\begin{equation}\label{eq:lennorm}
    \argmax_{\mathcal{C}_j\in\curchoices} \frac{\sum_{k=1}^{\ell_j}\log[P(\mathcal{C}_j^k| \mathbf{x}_i, \mathcal{C}_j^1\ldots \mathcal{C}_j^{k-1})]}{\ell_j}
\end{equation}
\subsection{Generalized Prompts}\label{sec:methods_generalizedprompts}
\begin{table*}[ht]
    \centering
    \begin{tabular}{l|p{.8\linewidth}}
        \toprule
         Task & Prompt \\
         \midrule
         Original & {\small\tt Sentence A: \textcolor{Blue}{\{\{sentence1\}\}} Sentence B: \textcolor{BrickRed}{\{\{sentence2\}\}}  "\textcolor{ForestGreen}{\{\{word\}\}}" has a similar meaning in sentences A and B. True or False?}\\
         Generalized & {\small\tt Sentence A: \textcolor{Blue}{\{\{premise\}\}} 
         Sentence B: \textcolor{BrickRed}{\{\{hypothesis\}\}} 
         "\textcolor{ForestGreen}{\{\{domain\}\}}" has a similar meaning in sentences A and B. 
         \textcolor{Orange}{\{\{ choice\_string \}\}}?}\\
        \midrule
        Example & {\small\tt Sentence A: \textcolor{Blue}{What is 2+2?} Sentence B: \textcolor{BrickRed}{Choices are: $\infty$, -10, fish, 4, $\sqrt{2}$} "\textcolor{ForestGreen}{math problem}" has a similar meaning in sentences A and B. \textcolor{Orange}{"A", "B", "C", "D" or "E"}?}\\
         \bottomrule
    \end{tabular}
    \caption{Sample prompt from WordsInContext Task \cite{wic-dataset} and its generalized form. Each {\tt \{\{ \}\}} represents an input from the dataset. The colors are the alignment of inputs.}
    \label{tab:example_prompts}
\end{table*}
We use the PromptSource\footnote{https://github.com/bigscience-workshop/promptsource} framework proposed by \citet{sanh2021multitask} for templates as it provides a standardized format for managing prompts. We standardize the input fields and answer formats across all tasks such that they fell into three general categories: {\sc Classification}, { \sc Entailment}, and {\sc Question Answering (QA)}. \autoref{tab:example_prompts} displays an example of what the generalization would look like for an {\sc Entailment} prompt. To use a prompt with a task that does not have the same number of inputs, we add additional task specific NL to better align the inputs. To use the example prompt from \autoref{tab:example_prompts} with a sentiment classification task, we would map the input text from the task to the {\tt premise } field and pass ``{\tt what is the sentiment}'' as the hypothesis. In prompts where answers choices are present, we replace them with an additional input field for the choice string(i.e. ``yes, no, or maybe'') to hold how the choices are presented constant across all prompts. A detailed breakdown of the tasks used for the generalized prompts can be found in \autoref{tab:generalprompttable}.
\section{Experimental Setup}\label{sec:experimentalsetup}
\subsection{Datasets}\label{sec:datasets}
For all datasets, we use the most recent version on HuggingFace \cite{wolf-etal-2020-transformers}. Every evaluation is done using the validation split as per \citet{sanh2021multitask}. For the evaluation datasets, we again follow \citet{sanh2021multitask} and use: Adversarial NLI (\textbf{ANLI}) \cite{anli-dataset}, CommitmentBank (\textbf{CB}) \cite{cb-dataset}, Recognizing Textural Entailment (\textbf{RTE}) \cite{rte-dataset-1,rte-dataset-2,rte-dataset-3,rte-dataset-4}, and Words In Context (\textbf{WiC}) \cite{wic-dataset}. \\
\indent Two additional tasks are used to evaluate T0's performance on complex tasks in unseen domains: \textbf{Algebra Question Answering (AQuA)} \cite{aqua-dataset} and \textbf{CraigslistBargains} \cite{he-etal-2018-decoupling}. Descriptions for these two tasks can be found in \autoref{sec:appendix_complex_tasks}.\\
\indent For the generalized prompts, we collect 86 prompts across 19 distinct tasks. In addition to these, we also include 12 prompts with no additional NL for each of the three categories for 4 different ablations. More details can be found in \autoref{sec:appendixgeneralprompts}.
\subsection{Model and Metrics}\label{sec:model}
We evaluate the performance of the 3B parameter T0, and T5 \cite{raffel-etal-2020} models with the HuggingFace implementation \cite{wolf-etal-2020-transformers} as we were limited to a single RTX Titan 24GB card. Following prior works \cite{sanh2021multitask,brown2020language,Holtzman2021SurfaceFC}, we report the accuracy and F1 scores on each of the eight datasets. 

As each task will have different mean metrics, we cannot compare the raw accuracy and F1 scores across tasks. For a given prompt, we calculate the median accuracy and F1 ranks compared to the 95 other prompts for all evaluation tasks. As the rank is ascending, \textbf{lower values} for median accuracy rank (MAR) and median F1 rank (MFR) indicate a better performing prompt.
\section{Results}\label{sec:results}

\subsection{Baselines On New Tasks}
\noindent As we evaluate T0-3B on two new tasks, we first want to gauge how the model performs as shown in both \autoref{fig:full_baselines} and \autoref{tab:medaccf1}. We follow \citet{sanh2021multitask} in using rank scoring without length normalization as defined by \autoref{eq:predprob} and find that for both AQuA and CraigslistBargains, the base T5 model performs better than T0. However, T0's unweighted multi-class F1 is better than that of T5 on both tasks, indicating that T5 achieves a higher score due to only predicting a subset of the choices. \autoref{fig:choices} provides further evidence for this hypothesis. In both AQuA and CraigslistBargains, T0 more evenly distributes its predictions across the possible choices where as T5 heavily favors a subset of the choices. Thus, the disparity in the accuracy is likely a result of class imbalance in the evaluation datasets in which T5 is 'lucky' in heavily predicting the class that was more populous. We consider this to be 'luck' as T5 outperforming T0 only occurs in only two of the datasets we examined. 
\subsection{Generalized Prompts}\label{sec:genpromptsresults}
\begin{table*}
    \centering
    \begin{adjustbox}{width=1\textwidth}
		\begin{tabular}{cl|cccccccc| c}
            \toprule
            {} & {} & ANLI R1 & ANLI R2 & ANLI R3 & AQuA & CB& Craigslist & RTE & WiC & Rank\\
            \midrule
            & No Prompt & 34.15 & 33.35 & 33.42 & 26.77 & 24.11 & 16.83 & 59.57 & 50.24 & 46.25\\
            \midrule
            \multirow{8}{2cm}{Unseen Prompts}& ANLI  & \textbf{37.60} & \cellcolor{green!25}\textbf{34.70} & \textbf{34.08} & 25.95 & \textbf{32.14} & 21.44 & 64.62 & \cellcolor{red!25}50.16 & 24.50\\
            & AQuA  & 36.10 & 33.40 & \cellcolor{green!25}35.42 & \cellcolor{red!25}\textbf{17.32} & \cellcolor{green!25}33.93 & 23.45 & 71.12 & 51.57 & \cellcolor{green!25}18.25\\
            & COPA & \cellcolor{green!25}39.30 & 34.40 & 34.00 & 20.47 & 26.79 & 16.58 & 69.31 & 50.63 & 21.25\\
            & Craigslist& \cellcolor{red!25}31.40 & \cellcolor{red!25}31.30 & 32.83 & 25.79 & \cellcolor{red!25}8.04 & \textbf{26.72} & \cellcolor{red!25}49.82 & 50.16 & \cellcolor{red!25}71.25\\
            & MathQA & 37.30 & 33.50 & 34.25 & 19.29 & 26.79 & 16.25 & \cellcolor{green!25}73.29 & 51.10 & 24.50\\
            & RTE& 36.10 & 33.20 & 33.58 & 22.05 & 23.21 & 20.27 & \textbf{61.37} & 50.47 & 43.25\\
            & SemEval2010 & 33.10 & 32.00 & 32.58 & \cellcolor{green!25}27.56 & 14.29 & 25.63 & 55.23 & 50.47 & 66.50\\
            & WiC & 31.75 & 33.45 & \cellcolor{red!25}32.33 & 26.57 & 13.39 & 18.01 & 55.05 & \textbf{50.47} & 64.25\\
            \midrule
            \multirow{3}{2cm}{Training Prompts}& AppReviews & 34.20 & 33.10 & 33.62 & 27.17 & 19.64 & \cellcolor{green!25}33.17 & 61.55 & 50.31 & 33.50\\
            & IMDB & 33.00 & 32.20 & 33.08 & 26.38 & 12.50 & \cellcolor{red!25}14.57 & 55.23 & 50.16 & \cellcolor{red!25}71.25\\
            & Yelp& 33.25 & 32.35 & 33.04 & 26.77 & 12.50 & 24.29 & 62.27 & \cellcolor{green!25}51.57 & 41.75\\
            \bottomrule
        \end{tabular}
    \end{adjustbox}
    \caption{Median Accuracy when using modified prompts for cross task zero-shot evaluation. \textbf{Bolded} entries are prompts for the original task. \colorbox{green!25}{\color{black} Green Cells} and \colorbox{red!25}{\color{black} Red Cells} are the best and worst performing tasks for a column respectively. Rank is the median rank of prompts from this task out of 95 total prompts. ANLI and CB both use the same prompts for their original task prompts per PromptSource. Some tasks are left out for clarity. The full table can be found in \autoref{table:full_crosstask}.} 
    \label{table:crosstask}
\end{table*} 
We report the results of the cross-task evaluation in \autoref{table:crosstask}. We find that the only task in which the original\footnote{By original we are referring to the prompts specifically designed for a task.} prompt performs best is ANLI R2 with an accuracy of 34.70. Conversely, the worst performing prompts for the AQuA task were its original prompts with an accuracy of 17.32. Furthermore, we find that there is no task out of the eight used for evaluation where not using a prompt has the best performance. 

We report how the added NL discussed in \autoref{sec:methods_generalizedprompts} effects the zero-shot performance in \autoref{tab:appendix_ablation_tables}. Across the eight evaluation tasks, adding some task specific text leads to an average increase of $4.65\%$ in the accuracy and a $2.34\%$ increase to the unweighted multiclass F1. However, there was also a decrease of $4.21\%$ and $13.90\%$ to minimum accuracy and unweighted multiclass F1 respectively. This implies that the added extra text helps to better amplify the negative and positive elements of a prompt. 
\subsection{Qualitative Analysis of Prompts}\label{sec:qualprompts}     
\autoref{tab:appendix_ablation_tables} displays the rank statistics across multiple ablations and \autoref{tab:appendix_correlations} displays the correlations of a prompt's qualities with their rank. In prompts which have choices, the MAR is 33.12 compared to 52.25 when the choices are left out. However, the range as indicated by the Q1 and Q3 MARs is significantly larger when the choices are included, implying that adding the choice string causes high variance.

We also find that when the choices are presented as multiple distinct choices\footnote{Presented with a clear delimiters/separation. For example A) yes B) no C) maybe} the MAR is further improves to 22.25. In comparison, the prompts with choices that are not in this format have a MAR of 36.00. These results provide further evidence to the findings from \citet{flan-paper} that clearly distinguishing the options in a prompt improves performance.

Next, we find that the median rank of prompts used in training is 50.25 compared to 42.00 of the unseen prompts. The F1 scores display a similar pattern with training prompts having a MFR of 55.75 while unseen prompts have a median of 36.50. Although this implies that prompts that share tokens with those used for training will perform worse, we find that there is no significant correlation between the number of tokens a prompt $\mathcal{P}$ shares with those used in training and its rank.

Finally, we find that longer prompts have a slightly negative impact on the performance of a prompt. \autoref{fig:appendix_ranks} shows that, with the 95 prompts used across eight tasks, the best performing prompts are those whose length is in the range $[14,21)$ as their MAR is 28.50 and MFR is 36.50. The Q1 values are 18.00 and 15.38, respectively. In comparison, we find that prompts with lengths with lengths $\geq 25$ have a median MAR of 50.25 and MFR of 72.00, indicating a negative impact on performance. Surprisingly, we find that prompts whose lengths are $< 14$ have a median MAR of 47.75 and MFR of 49.00. While this is a negative impact on performance, it is not as large as that in longer prompts.

\section{Related Works}\label{sec:related}
\textbf{Pre-trained Language Models} In the past few years, large pre-trained models have rose to prominence due to their strong performance on a wide range of NLP tasks \cite{Radford2019LanguageMA,brown2020language,lewis-etal-2020-bart}. In particular, T5 \cite{raffel-etal-2020} explored transferred learning for large LMs by transforming all NLP problems to a text-to-text format. Beyond natural language, these models have also excelled at tasks in other modalities such as code generation \cite{Austin2021ProgramSW,Chen2021EvaluatingLL,orlanski-gittens-2021-reading}, in-context learning \cite{Min2021MetaICLLT, Zhong2021AdaptingLM}, and semantic parsing \cite{Rongali2020DontPG}. One aspect of large LMs is that they perform well in zero-shot settings \cite{Radford2019LanguageMA,brown2020language, vu-etal-2020-exploring}. 

\noindent\textbf{NL Prompting} A drawback of these large LMs is that their size makes it costly to fine-tune them. This lead to the rise of the ``\textit{pre-train, prompt, and predict}'' paradigm in which a downstream tasks are modified to resemble those used in training through the use of NL prompts \cite{Liu2021PretrainPA}. These prompts have improved few-shot and zero-shot performance across a vast number of models and tasks \cite{brown2020language,schick-schutze-2021-just,schick-schutze-2021-exploiting,Mishra2021CrossTaskGV,Scao2021HowMD,shin-etal-2020-autoprompt}. Recent models such as FLAN \cite{wei2021finetuned} and T0 \cite{sanh2021multitask} have shown that even better zero-shot performance can be achieved through using a multi-task pre-training objectives with a diverse set of prompts.

\section{Conclusion}

In this paper, we examined T0's performance on a range of fixed multiple-choice tasks. We find that T0 does worse than T5 on two unseen complex tasks. Next, we evaluated how the performance of a prompt transfers between tasks. Our results show that using a prompt performs consistently better than not using any prompt. We conclude with a quantitative analysis of what aspects of a prompt affect its performance. We find that prompts with the choices in them are 66.82\% better than those that leave the choices out. Next, we find that prompts not used in pre-training are 19.64\% better than those that were. Finally, we find that prompts whose length are between 14 and 24 perform better than both longer and shorter prompts. Further work should examine prompt transfer with larger models while also expanding the number of prompts and tasks used.


\bibliographystyle{acl_natbib}
\bibliography{references}

\appendix
\section{Generalized Prompts}\label{sec:appendixgeneralprompts}
The tasks we took prompts from that were not used to train T0 are:
\begin{itemize}
    \item COPA \cite{copa-dataset}
    \item FinancialNews \cite{financial-phrasebank-dataset}
    \item LAMBADA \cite{lambada-dataset}
    \item MathQA \cite{math-qa-dataset}
    \item MultiXSci \cite{multi-sci-sum-dataset} 
    \item NumerSense \cite{numersense-dataset}
    \item SemEval2010 \cite{semeval-dataset}
    \item ZEST \cite{zest-dataset}
\end{itemize}
The tasks we took prompts from that \textit{were} used to train T0 are:
\begin{itemize}
    \item AppReviews \cite{app-reviews-dataset}
    \item Adversarial QA \cite{adversarial-qa-dataset}
    \item CommonGen \cite{commongen-dataset}
    \item IMDB \cite{imdb-dataset} 
    \item XSum \cite{xsum-dataset}
\end{itemize}
\begin{table}[H]
    \centering
    \begin{tabular}{l|lll}
            \toprule
            Task Name & \# & Type & MCQ\\
            \midrule
            Text & 6 & All Three &  \ding{51} \\
            Text+Choices & 6 & All Three &   \ding{51}\\
            \midrule
            ANLI  & 9 & {\sc Entailment} & \\
            AQuA  & 5 & {\sc Classification} & \ding{51}\\
            COPA & 3 & {\sc Entailment} & \\
            Craigslist & 4 & {\sc Classification} & \ding{51} \\
            FinancialNews &4 & {\sc Classification} &\\
            LAMBADA & 3 & {\sc Classification} & \\
            MathQA & 5 & {\sc Classification} & \ding{51} \\
            MultiXSci & 4 & {\sc Classification} & \\
            NumerSense & 5 & {\sc Classification} &\\
            RTE& 5 & {\sc Entailment} &\\
            SemEval2010 & 3 & {\sc Classification}&\\
            WiC & 4 & {\sc Entailment} &\\
            ZEST & 4 & {\sc QA}\\
            \midrule
            AppReviews & 2 & {\sc Classification} & \\
            AdversarialQA & 4 & {\sc QA} &\\
            CommonGen & 2 & {\sc Classification} &\\
            IMDB & 5 & {\sc Classification} &\\
            XSum & 4 & {\sc Classification} &\\
            Yelp & 4 & {\sc Classification} &\\
            \bottomrule
        \end{tabular}
    \caption{Number of prompts used by task for the generalized prompts. \# is the number of prompts used from this task. MCQ indicates if the task had prompts that are formatted as an multiple choice question with choice letters.}
    \label{tab:generalprompttable}
\end{table}
\section{Complex Task Datasets}\label{sec:appendix_complex_tasks}
The two complex task used to evaluate T0's performance are:

\noindent \textbf{Algebra Question Answering (AQuA)} Dataset of multiple choice algebraic word problems. The choices for this task are \{{\sc A,B,C,D,E}\} and each letter maps to a potential mathematical answer \cite{aqua-dataset}.\\
\noindent\textbf{CraigslistBargains} A collection of dialogues involving two-parties negotiating the price of an item for sale on Craigslist. For the scope of this paper, we use the task of classifying who won the negotiation \cite{he-etal-2018-decoupling}.\\
\section{Additional Results}\label{sec:rawresults}
This section is for results that could not be included in the main body of the paper due to the page limits.
\begin{figure*}
    \centering
    \includegraphics[width=1\textwidth]{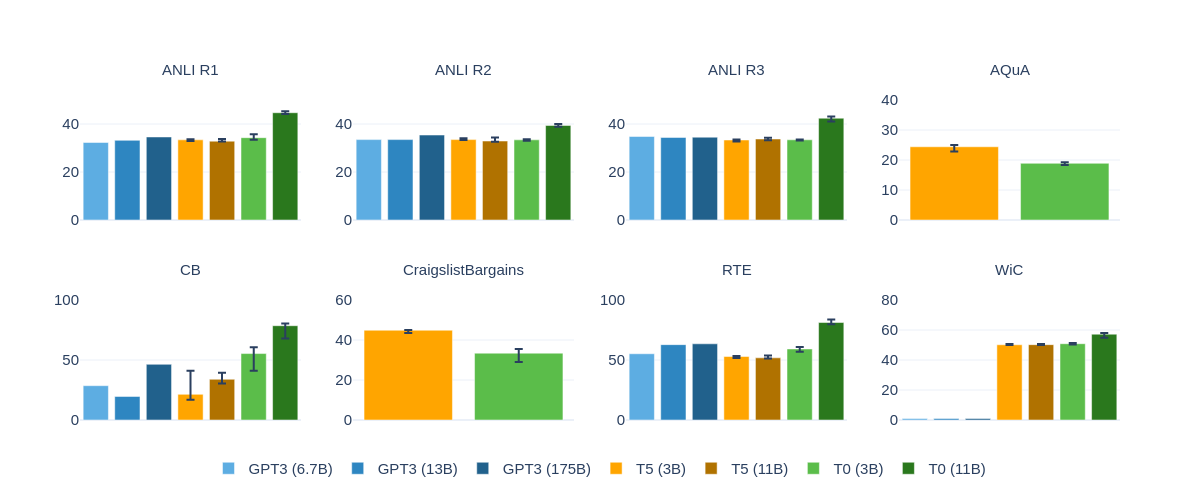}
    \caption{Median Accuracy with interquartile range for three models:\colorbox{GPTBlue}{\color{black} \textbf{GPT-3}}, \colorbox{t5orange}{\color{black} \textbf{T5}}, and \colorbox{t0green}{\color{black} \textbf{T0}}. Darker indicates larger model. Results for GPT-3 Model are from \citet{brown2020language}. Results for the 11B T0 and T5 models are taken from \citet{sanh2021multitask}}
    \label{fig:full_baselines}
\end{figure*}

\begin{table}[H]
    \centering
    \begin{tabular}{ll|cc}
        \toprule
        Task & Model & Accuracy & F1\\
        \midrule
        \multirow{2}{3cm}{ANLI R1} & T0 &     34.30 & 26.17 \\
        & T5 &     33.40 & 20.28 \\
        \hline
        \multirow{2}{3cm}{ANLI R2} & T0 &     33.40 & 23.70 \\
        & T5 &     33.50 & 21.25 \\
        \hline
        \multirow{2}{3cm}{ANLI R3} & T0 &     33.42 & 21.82 \\
        & T5 &     33.33 & 24.84 \\
        \hline
        \multirow{2}{3cm}{AQuA} & T0 &     18.90 & 15.04 \\
        & T5 &     24.41 & 12.74 \\
        \hline
        \multirow{2}{3cm}{CB} & T0 &     55.36 & 38.62 \\
        & T5 &     21.43 & 19.41 \\
        \hline
        \multirow{2}{3cm}{CraigslistBargains} & T0 &     33.42 & 18.45 \\
        & T5 &     44.89 & 16.47 \\
        \hline
        \multirow{2}{3cm}{RTE} & T0 &     59.21 & 69.56 \\
        & T5 &     52.89 & 36.08 \\
        \hline
        \multirow{2}{3cm}{WiC} & T0 &     50.86 &  8.81 \\
        & T5 &     50.16 &  5.44 \\
        \bottomrule
    \end{tabular}
    \caption{Median accuracy and F1 for the corresponding \autoref{fig:full_baselines}.}
    \label{tab:medaccf1}
\end{table}
\begin{figure}[h]
    \centering
    \includegraphics[scale=1.5,width=\linewidth]{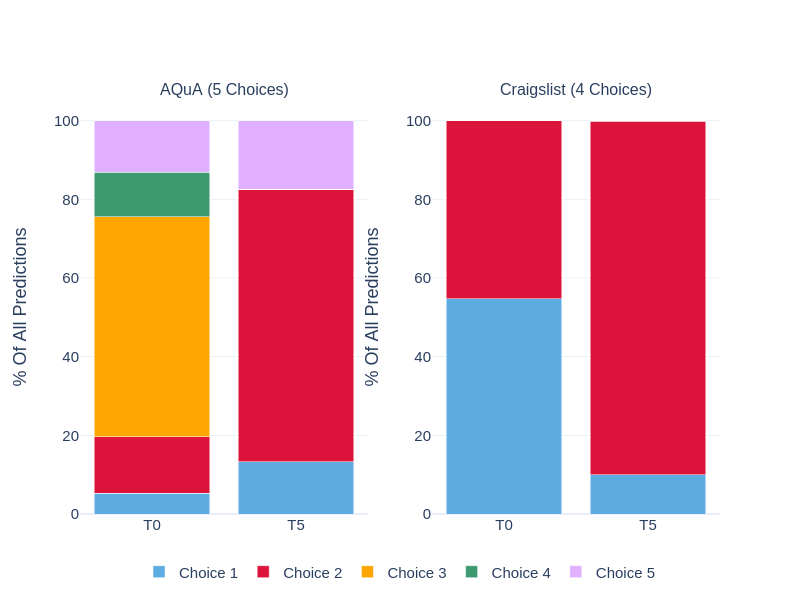}
    \caption{Distribution of choices for T0 and T5 on the AQuA and CraigslistBargains. }
    \label{fig:choices}
\end{figure}
\begin{table*}
    \centering
    \begin{tabular}{l|rrrr|rrrr}
        \toprule
        {} & \multicolumn{4}{c|}{Accuracy} &\multicolumn{4}{c}{F1} \\
        Ablation &   Mean &  Median &  Q1 &  Q3 &  Mean &   Median &  Q1 &  Q3 \\
        \midrule
        Training Prompts &          51.46 &            50.25 &        45.25 &        63.00 &    54.18 &      55.75 &  38.00 &  66.00 \\
        Unseen Prompts   &          42.72 &            42.00 &        23.50 &        60.75 &    42.46 &      36.50 &  22.00 &  62.50 \\
        \midrule
        With Choices     &          39.44 &            33.12 &        20.19 &        58.62 &    39.37 &      31.00 &  19.12 &  61.75 \\
        No Choices       &          51.73 &            52.25 &        44.75 &        60.50 &    55.93 &      53.50 &  43.00 &  66.00 \\
        \midrule
        Is MCQ           &          25.80 &            22.25 &        16.50 &        26.00 &    23.14 &      16.50 &  13.75 &  25.25 \\
        Not MCQ          &          43.28 &            36.00 &        26.38 &        62.62 &    43.95 &      36.50 &  22.00 &  65.50 \\
        \midrule
        Extra Text       &          44.99 &            46.75 &        28.81 &        60.31 &    46.44 &      46.50 &  27.75 &  66.00 \\
        No Extra Text    &          44.41 &            48.00 &        32.75 &        57.62 &    45.43 &      44.50 &  26.62 &  62.25 \\
        \bottomrule
        \end{tabular}
    \caption{Accuracy and F1 ranks for different ablations. It is calculated by taking the median rank of a given prompt across all 8 tasks then taking the Mean, Median, Q1, and Q3 of that. Lower is better. Q1 and Q3 are the first and third quartile.}
    \label{tab:appendix_ablation_tables}
\end{table*}
\begin{table*}
    \centering
    \begin{adjustbox}{width=1\textwidth}
		\begin{tabular}{cl|cccccccc| c}
            \toprule
            {} & {} & ANLI R1 & ANLI R2 & ANLI R3 & AQuA & CB& Craigslist & RTE & WiC & Rank\\
            \midrule
            & No Prompt & 34.15 & 33.35 & 33.42 & 26.77 & 24.11 & 16.83 & 59.57 & 50.24 & 46.25\\
            \midrule
            \multirow{13}{2cm}{Unseen Prompts}& ANLI  & \textbf{37.60} & \cellcolor{green!25}\textbf{34.70} & \textbf{34.08} & 25.95 & \textbf{32.14} & 21.44 & 64.62 & \cellcolor{red!25}50.16 & 24.50\\
            & AQuA  & 36.10 & 33.40 & \cellcolor{green!25}35.42 & \cellcolor{red!25}\textbf{17.32} & \cellcolor{green!25}33.93 & 23.45 & 71.12 & 51.57 & \cellcolor{green!25}18.25\\
            & COPA & \cellcolor{green!25}39.30 & 34.40 & 34.00 & 20.47 & 26.79 & 16.58 & 69.31 & 50.63 & 21.25\\
            & Craigslist& \cellcolor{red!25}31.40 & \cellcolor{red!25}31.30 & 32.83 & 25.79 & \cellcolor{red!25}8.04 & \textbf{26.72} & \cellcolor{red!25}49.82 & 50.16 & \cellcolor{red!25}71.25\\
            & FinNews & 33.05 & 31.65 & 32.83 & 25.95 & 18.75 & 19.68 & 55.78 & 50.31 & 64.00\\
            & LAMBADA & 34.00 & 32.40 & 32.50 & 26.77 & 19.64 & 16.08 & 57.76 & 50.78 & 58.50\\
            & MathQA & 37.30 & 33.50 & 34.25 & 19.29 & 26.79 & 16.25 & \cellcolor{green!25}73.29 & 51.10 & 24.50\\
            & Multi-XSci & 34.20 & 32.70 & 32.75 & 27.17 & 19.64 & 19.43 & 58.84 & 50.31 & 54.75\\
            & NumerSense & 37.70 & 33.30 & 33.17 & 25.20 & 25.00 & 15.75 & 65.70 & 50.63 & 40.50\\
            & RTE& 36.10 & 33.20 & 33.58 & 22.05 & 23.21 & 20.27 & \textbf{61.37} & 50.47 & 43.25\\
            & SemEval2010 & 33.10 & 32.00 & 32.58 & \cellcolor{green!25}27.56 & 14.29 & 25.63 & 55.23 & 50.47 & 66.50\\
            & WiC & 31.75 & 33.45 & \cellcolor{red!25}32.33 & 26.57 & 13.39 & 18.01 & 55.05 & \textbf{50.47} & 64.25\\
            & ZEST & 35.20 & 32.65 & 33.38 & 26.77 & 23.21 & 17.76 & 66.79 & 50.71 & 38.25\\
            \midrule
            \multirow{5}{2cm}{Training Prompts}& AppReviews & 34.20 & 33.10 & 33.62 & 27.17 & 19.64 & \cellcolor{green!25}33.17 & 61.55 & 50.31 & 33.50\\
            & CommonGen & 33.75 & 33.35 & 32.50 & 25.39 & 13.39 & 23.62 & 51.81 & 51.18 & 58.75\\
            & IMDB & 33.00 & 32.20 & 33.08 & 26.38 & 12.50 & \cellcolor{red!25}14.57 & 55.23 & 50.16 & \cellcolor{red!25}71.25\\
            & XSum & 33.50 & 32.00 & 33.00 & 26.97 & 10.71 & 19.26 & 57.22 & 50.86 & 58.50\\
            & Yelp& 33.25 & 32.35 & 33.04 & 26.77 & 12.50 & 24.29 & 62.27 & \cellcolor{green!25}51.57 & 41.75\\
            \bottomrule
        \end{tabular}
    \end{adjustbox}
    \caption{Median Accuracy when using modified prompts for cross task zero-shot evaluation. \textbf{Bolded} entries are prompts for the original task. \colorbox{green!25}{\color{black} Green Cells} and \colorbox{red!25}{\color{black} Red Cells} are the best and worst performing tasks for a column respectively. Rank is the median rank of prompts from this task out of 95 total prompts. ANLI and CB both use the same prompts for their original task prompts per PromptSource.} 
    \label{table:full_crosstask}
\end{table*} 
\begin{table}[H]
    \centering
    \begin{tabular}{l|rr}
        \toprule
        {} &  Accuracy &  F1 \\
        \midrule
        Has Choices &          -0.14 &    -0.27 \\
        Is MCQ            &          -0.16 &    -0.25 \\
        Training Prompt     &           0.07 &     0.13 \\
        Length    &           0.14 &     0.18 \\
        \bottomrule
        \end{tabular}
    \caption{Correlations with metric rank for a given prompt quality. Per the definition of rank, a lower score is better and therefore a negative correlation indicates a quality improves performance. Length is measured as the raw number of tokens in a prompt.}
    \label{tab:appendix_correlations}
\end{table}

\begin{figure*}[ht]
    \centering
    \includegraphics[width=\linewidth,keepaspectratio]{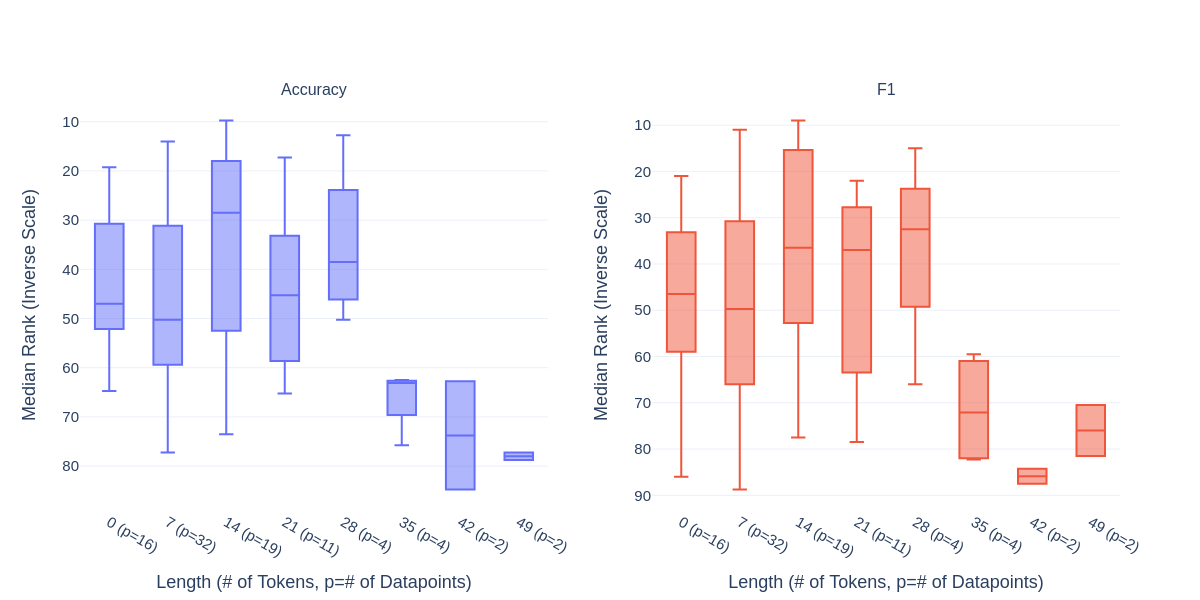}
    \caption{Accuracy and F1 rank compared with the number of tokens in the prompt. The tick value is the lower bound of the range. p=The number of prompts that fall into that respective range. }
    \label{fig:appendix_ranks}
\end{figure*}

\end{document}